\documentclass[lettersize,journal]{IEEEtran}
\usepackage[table]{xcolor}
\usepackage{amsmath,amsfonts}
\usepackage{array}
\usepackage{textcomp}
\usepackage{stfloats}
\usepackage{url}
\usepackage{verbatim}
\usepackage{graphicx}
\usepackage{cite}
\usepackage{makecell}
\usepackage{microtype}
\usepackage{graphicx}
\usepackage{subcaption}
\usepackage{booktabs} 
\usepackage{times}
\usepackage{epsfig}
\usepackage{graphicx}
\usepackage{caption}
\usepackage{amsmath}
\usepackage{amssymb}
\usepackage{booktabs}
\usepackage{multirow}
\usepackage{enumitem}
\usepackage{diagbox} 
\usepackage{array} 
\usepackage{hyperref}
\begin{document}

\title{InterCoG: Towards Spatially Precise Image Editing with Interleaved Chain-of-Grounding Reasoning}

\author{Yecong Wan, Fan Li, Chunwei Wang, Hao Wu, Mingwen Shao~\IEEEmembership{Member,~IEEE}, and Wangmeng Zuo~\IEEEmembership{Senior Member,~IEEE}
	\thanks{Yecong Wan, Hao Wu, and Wangmeng Zuo are with the Faculty of Computing, Harbin Institute of Technology, Harbin, 150001, China. Yecong Wan is also with the Zhengzhou Advanced Research Institute of Harbin Institute of Technology, Zhengzhou, 450000, China.}
	\thanks{Fan Li and Chunwei Wang are with the Huawei Noah’s Ark Lab, Shenzhen, 518100, China.}
	\thanks{Mingwen Shao is with the Artificial Intelligence Research Institute, Shenzhen University of Advanced Technology, Shenzhen, 518107, China.}

}
\markboth{Journal of \LaTeX\ Class Files,~Vol.~14, No.~8, August~2021}%
{Shell \MakeLowercase{\textit{et al.}}: A Sample Article Using IEEEtran.cls for IEEE Journals}


\maketitle

\begin{abstract}
Emerging unified editing models have demonstrated strong capabilities in general object editing tasks. However, it remains a significant challenge to perform fine-grained editing in complex multi-entity scenes, particularly those where targets are not visually salient and require spatial reasoning. To this end, we propose InterCoG, a novel text-vision Interleaved Chain-of-Grounding reasoning framework for fine-grained image editing in complex real-world scenes.
The key insight of InterCoG is to first perform object position reasoning solely within text that includes spatial relation details to explicitly deduce the location and identity of the edited target. It then conducts visual grounding via highlighting the editing targets with generated bounding boxes and masks in pixel space, and finally rewrites the editing description to specify the intended outcomes. 
To further facilitate this paradigm, we propose two auxiliary training modules: multimodal grounding reconstruction supervision and multimodal grounding reasoning alignment to enforce spatial localization accuracy and reasoning interpretability, respectively. We also construct GroundEdit-45K, a dataset comprising 45K grounding-oriented editing samples with detailed reasoning annotations, and GroundEdit-Bench, a manually curated benchmark for grounding-aware editing evaluation. Extensive experiments substantiate the superiority of our approach in highly precise edits under spatially intricate and multi-entity scenes. The code and dataset will be publicly available.
\end{abstract}


\section{Introduction}
\vspace{-3pt}



Instruction-based image editing has emerged as a fundamental paradigm for controllable visual content manipulation, enabling users to modify images through natural language instructions (e.g., object addition, removal, attribute modification)~\cite{brooks2023instructpix2pix,zhang2023magicbrush,yu2025anyedit,ge2024seed,he2025freeedit,xia2025consistent,feng2025instruction}. 
Recent advances have significantly improved the fidelity and diversity of generated results \cite{yu2025anyedit,wu2025qwen,deng2025emerging,liu2025step1x}, moving the field closer to practical deployment. 

Despite this progress, a fundamental challenge remains largely unresolved: \emph{how to precisely identify and manipulate the intended editing target in complex scenes where the target is implicitly specified and not visually salient}. 
Unlike simple scenarios where the target is explicitly mentioned (e.g., “the red car”), real-world editing instructions often involve referential and relational reasoning, such as “the person standing behind the table” or “the second object from the left near the window.” 
These cases require not only understanding the semantic intent, but also inferring spatial relationships and resolving ambiguity among multiple similar entities, posing a significant challenge for existing methods.

Early diffusion-based approaches~\cite{li2024brushedit,brooks2023instructpix2pix,zhang2023magicbrush,batifol2025flux} typically rely on pretrained text encoders such as CLIP~\cite{radford2021learning} or T5~\cite{raffel2020exploring} to guide image manipulation. 
While effective for coarse and globally aligned edits, these methods inherently lack explicit reasoning mechanisms and fine-grained grounding capabilities \cite{huang2024smartedit}, making them prone to ambiguous or imprecise edits in multi-entity scenes.

More recently, unified multimodal generation frameworks~\cite{wu2025qwen,liu2025step1x,lin2025uniworld,chen2025blip3,deng2025emerging,pan2025transfer,wang2025ovis,openai2025gpt4o} have incorporated large multimodal language models (MLLMs) \cite{bai2025qwen2} as reasoning-capable text encoders, demonstrating remarkable improvements in instruction following and semantic consistency. 
However, their editing capability still predominantly relies on implicit alignment between text and image features \cite{lin2025uniworld,liu2025step1x}, which tends to favor salient or explicitly described targets. 
When faced with compositional scenes containing multiple similar entities, these models often fail to disambiguate the correct target, leading to erroneous or entangled edits. In addition, recent efforts have explored region-aware editing via pre-extracted candidate regions~\cite{zhou2025fireedit}, explicit coordinate prediction~\cite{fang2025got,qu2025replan}, or coarse mask-based localization~\cite{zou2025beyond}. However, these explicit geometric cues often prove coarse and unreliable, failing to satisfy the demands of multi-subject scenes that require implicit, reasoning-based grounding cues for the target editing regions.

To address the aforementioned challenges, we propose \textbf{InterCoG}, a novel text–vision \textbf{Inter}leaved \textbf{C}hain-\textbf{o}f-\textbf{G}rounding framework that enables precise editing for subjects and regions based on the context of implicit reasoning. Specifically, InterCoG first performs object position reasoning solely within the text modality; that is, it deduces spatial relation details to clarify the location and identity of the edited targets. Following this, InterCoG performs visual grounding by highlighting the targets with generated bounding boxes and masks in the pixel space, and finally rewrites the editing description to specify the intended outcomes.
By interleaving these three fine-grained textual and visual localization cues, our method establishes a specific reasoning trajectory that accurately anchors editable areas, ensuring that edits remain semantically faithful and spatially consistent. 
To further facilitate this paradigm, we introduce two auxiliary training modules: multimodal grounding reconstruction supervision and multimodal grounding reasoning alignment. The former enables the model to enhance its object grounding capabilities via mask decoding, while the latter explicitly aligns the features of the target regions with text–vision grounding cues. This thereby enhances the interpretability and spatial alignment of the reasoning process.

To facilitate grounding-oriented and fine-grained editing, we construct GroundEdit-45K, which contains 45K editing samples. Each sample is annotated with detailed text–vision chain-of-grounding annotations, where every editing target is provided with precise bounding-box and mask tags. 
Furthermore, we introduce GroundEdit-Bench to evaluate models’ capabilities in complex, grounding-aware editing. Comprehensive experiments substantiate the superiority of InterCoG over alternatives under a variety of real-world challenging conditions.

In conclusion, the main contributions are as follows.
\begin{itemize}[noitemsep]
	\item We propose InterCoG, an innovative interleaved chain-of-grounding reasoning framework that enables fine-grained editing in semantically dense and compositionally complex scenes.
	\item We introduce two auxiliary training modules - multimodal grounding reconstruction supervision and multimodal grounding reasoning alignment - to enhance spatial localization accuracy and reasoning interpretability. 
	\item We construct GroundEdit-45K, comprising 45K samples with explicit chain-of-grounding annotations, and GroundEdit-Bench for benchmarking grounding-aware editing.
	\item Our InterCoG demonstrates significant improvements in challenging grounding-oriented editing, substantially outperforming existing methods across various real-world complex scenarios.
\end{itemize}

\begin{figure*}[t]
	\vspace{-8pt}
	\begin{center}
		\includegraphics[width=\linewidth]{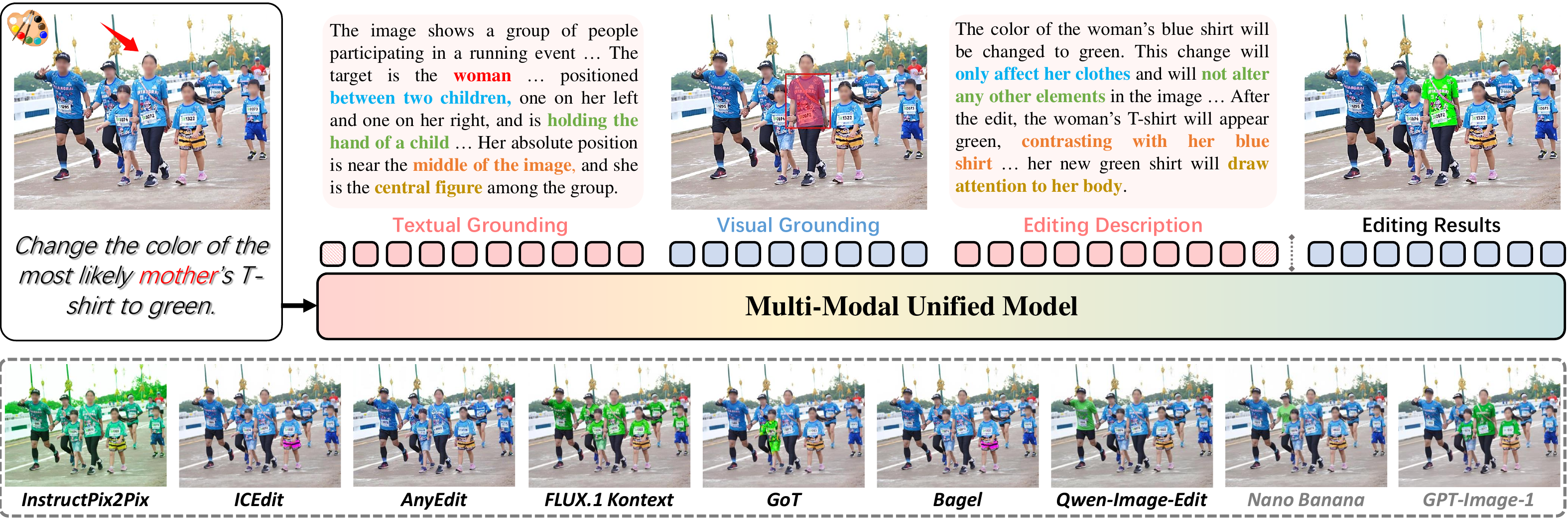}
	\end{center}
	\vspace{-8pt}
	\caption{ We present \textbf{InterCoG}, a novel interleaved chain-of-grounding reasoning framework capable of performing spatially precise image editing in complex scenes, particularly those demanding referential reasoning for precise target localization. InterCoG first performs textual grounding to parse the user-specified editing target, followed by visual grounding with explicit and interpretable localization cues in the form of mask and bounding boxe tags. Leveraging these grounded cues, the model is able to conduct target-faithful editing under complex multi-entity scenarios, thus enjoying better functionality and practicality in realistic applications.}
	\label{figure1}
	\vspace{-9pt}
\end{figure*}

\section{Related Work}
\subsection{Instruction-based Image Editing}
\vspace{-1pt}
Instruction-based image editing demands models to interpret nuanced user intentions and execute edits with high semantic fidelity and spatial precision. Pioneering works such as InstructPix2Pix \cite{brooks2023instructpix2pix}, MagicBrush \cite{zhang2023magicbrush}, and BrushEdit \cite{li2024brushedit} learn paired instruction–edit mappings with diffusion models but lack explicit instruction representation learning, resulting in shallow semantic understanding and limited alignment with fine-grained user intent. Subsequent efforts including AnyEdit~\cite{yu2025anyedit}, UltraEdit~\cite{zhao2024ultraedit}, EmuEdit~\cite{sheynin2024emu}, and ICEdit~\cite{zhang2025context} extend this paradigm through larger datasets and refined architectures, yet they remain limited to handling complex instructions and intricate image compositions. 
The recent emergence of GPT-4o~\cite{openai2025gpt4o} underscores the advantages of unifying visual understanding and generation. Parallel efforts either harness the instruction comprehension capacity of MLLMs as control interfaces for generative models~\cite{lin2025uniworld,wu2025omnigen2,wu2025qwen,chen2025blip3,ge2024seed} or pursue intrinsically unified multimodal architectures~\cite{deng2025emerging,xie2024show,ma2025janusflow}, collectively advancing the fidelity and controllability of instruction-driven editing and ushering in semantically coherent visual generation. Despite this progress, current models typically excel at identifying “what to edit” but often fail to reason “where to edit”, leading to suboptimal target grounding and degraded editing quality in complex multi-subjects scenes.

\vspace{-2pt}
\subsection{Textual and Visual Chain of Thought }
\vspace{-2pt}

Large Language Models (LLMs) have exhibited exceptional reasoning and comprehension abilities through chain-of-thought (CoT) reasoning~\cite{wei2022chain}, enabling the decomposition of complex tasks into transparent intermediate steps. A surge of recent studies~\cite{mitra2024compositional,zhang2023multimodal,zheng2023ddcot,gupta2023visual,suris2023vipergpt,hu2024visual,openai2025thinking} extend CoT reasoning into the multimodal domain by introducing explicit visual operations, enabling MLLMs to perform coherent vision–language reasoning with interpretable spatial manipulation, such as cropped, rotated, or zoomed views. 
Beyond question answering, GoT~\cite{fang2025got} exploits MLLMs to jointly reason about spatial semantics and scene layout, while ReasonBrain~\cite{he2025reasoning} repurposes MLLMs for reasoning-driven editing via reasoning cue extraction and cross-modal enhancement. Draw-In-Mind~\cite{zeng2025draw} and EditThinker~\cite{li2025editthinker} further re-plans editing operations by reformulating the instruction through imagination-based reasoning.
With the rapid evolution of unified multimodal models, interleaved text–vision reasoning has emerged~\cite{li2025zebra,shi2025mathcanvas,qin2025uni,zou2025beyond,huang2025interleaving}.
For instance, UniCoT~\cite{qin2025uni} performs iterative self-reflection to refine generation quality. MURE~\cite{zou2025beyond} integrates interleaved multimodal generation for image editing, yet still struggles with referential reasoning and pixel-level grounding in compositional complex scenes.
Moreover, most existing frameworks lack explicit mechanisms to align intermediate reasoning with final outcomes, leading to reasoning–output inconsistencies that underexploit the full potential of CoT reasoning.
\begin{figure*}[t]
	\begin{center}
		\includegraphics[width=\linewidth]{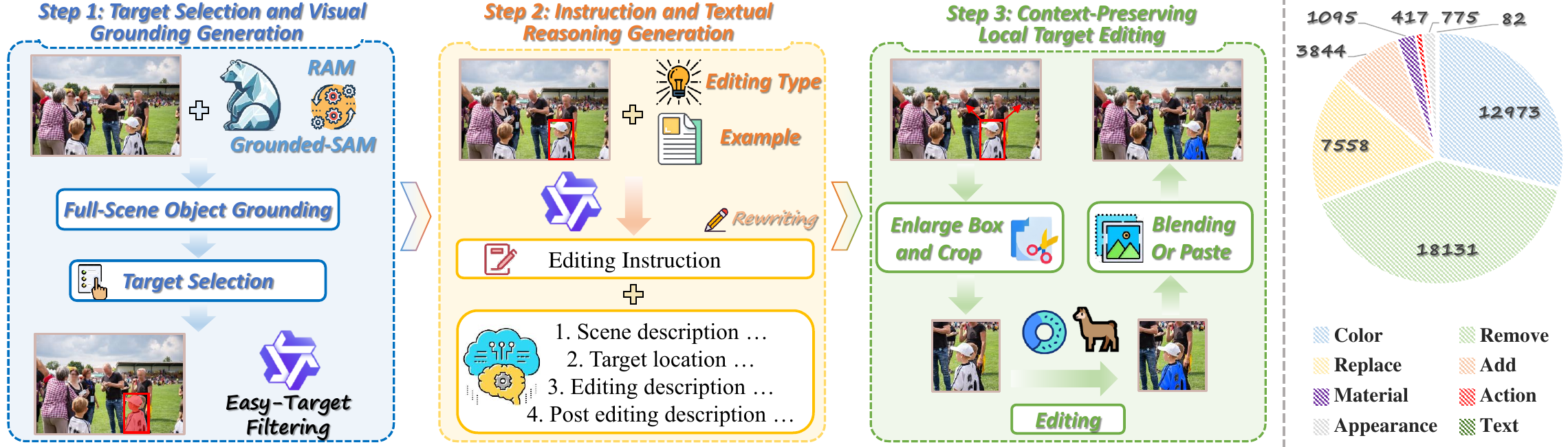}
	\end{center}
	\vspace{-8pt}
	\caption{\textbf{GroundEdit-45K Dataset Construction Pipeline and Statistics.} \textbf{Left:} The pipeline consists of three steps: (1) target selection and visual grounding generation; (2) instruction and text reasoning generation; and (3) context-preserving local target editing. \textbf{Right:} Our dataset contains 45K samples covering 8 categories of local editing types.}
	\label{figure2}
	\vspace{-10pt}
\end{figure*}

\section{Methodology}

Editing real-world images often entails scenes that comprise multiple entities, overlapping elements, and complex spatial arrangements.
However, despite the remarkable visual quality achieved by recent image editing methods \cite{lin2025uniworld,wu2025omnigen2,wu2025qwen}, they typically struggle to accurately localize the referential targets under such challenging conditions. To address these challenges, we introduce InterCoG, a text–vision interleaved chain-of-grounding paradigm that alternately exploits the logical reasoning ability of text and the fine-grained spatial grounding capability of vision, enabling spatially precise and interpretable editing in complex real-world scenarios.

As illustrated in Fig.~\ref{figure1}, the interleaved chain-of-grounding paradigm unfolds through four major steps: 
\textbf{(1)} The model first analyzes the input image and instruction to infer the user-intended editing target, producing a textual-level scene understanding and position description;
\textbf{(2)} Guided by this reasoning, it performs visual grounding by generating an image annotated with a red bounding box and a semi-transparent mask, precisely perceiving and localizing the desired targets;
\textbf{(3)} The model subsequently constructs a fine-grained editing description accompanied by an outcome rationale; and
\textbf{(4)} the interleaved reasoning chain ultimately directs the model to execute final editing.

By interleaving linguistic reasoning and spatial grounding, InterCoG establishes a coherent bridge between referential comprehension and localized visual manipulation, achieving fine-grained, semantically faithful edits across complex multi-entity scenarios.
We next introduce the key elements of our paradigm: dataset construction and framework.

\subsection{Grounded Editing Dataset Construction}
Existing image editing datasets primarily focus on salient objects or prominent regions, offering limited coverage of fine-grained and contextually entangled targets. To bridge this gap and support our proposed chain-of-grounding paradigm, we construct GroundEdit-45K, a meticulously crafted dataset comprising 45K grounded editing samples, each equipped with natural language instructions, text–vision interleaved chain-of-grounding annotations, and corresponding images.

As illustrated in Fig. \ref{figure2}, we adopt SAM-1B \cite{kirillov2023segment} as the source of raw images, given its rich diversity in spatial layouts, multi-entity compositions, and large-scale real-world scenes.
In contrast to prior datasets such as ImgEdit~\cite{ye2025imgedit}, which first define editing targets and then perform grounding, our setting focuses on complex multi-entity scenarios in which existing grounding models~\cite{lai2024lisa,ren2024grounded} often fail to accurately localize the desired targets. To ensure precise grounding annotations, we adopt a grounding-first strategy: the entire scene is first grounded, after which specific editing targets are selected from the grounded entities to prompt MLLMs to generate the corresponding editing instructions. 

\noindent\textbf{Step 1.} We begin by identifying all potential objects using the Recognize Anything Model (RAM)~\cite{huang2025open}, followed by Grounded-SAM~\cite{ren2024grounded} to localize all detected entities across the image, yielding corresponding bounding boxes and segmentation masks. We filter out low-confidence predictions and randomly select one remaining object as the editing target.
To ensure that the selected targets exhibit non-trivial grounding difficulty rather than trivial recognition, we employ Qwen2.5-VL-7B~\cite{bai2025qwen2} to filter out overly salient instances.
The final visual grounding is then defined as the grounded image itself, with the target highlighted by a red semi-transparent mask (opacity 0.5) and a red bounding box.

\noindent\textbf{Step 2.} We overlay the bounding box of the selected target onto the original image and prompt Qwen2.5-VL-72B~\cite{bai2025qwen2} with predefined editing types to generate the editing instruction together with its structured textual reasoning.
Each reasoning instance consists of four components:
(1) scene description, (2) target localization, (3) editing description, and (4) post editing description.
Among them, (1) and (2) constitute the first reasoning stage, focusing on referential and spatial comprehension, while (3) and (4) form the second reasoning stage, describing the editing intent and the expected visual outcome.
To encourage reasoning-aware instruction generation, we prompt the model to produce instructions such as \textit{“Turn the clothes of the person facing to the left to blue.”}, where identifying the correct target requires resolving multi-entity spatial relations rather than trivial object naming.
Qwen3~\cite{yang2025qwen3} is further employed to rewrite and paraphrase the generated instructions, yielding more natural and varied expressions.

\noindent\textbf{Step 3.} To guarantee that the designated target is accurately modified, we first crop the region specified by its bounding box and apply Bagel~\cite{deng2025emerging} for localized editing.
The edited patch is then reintegrated into the original image via blending, ensuring natural boundary transitions and overall visual consistency.
For removal tasks, we employ the target mask with LaMa~\cite{suvorov2022resolution} to synthesize content-consistent inpainting results, whereas for other editing types, Bagel is used to generate the corresponding modifications.
Finally, all edited samples undergo quality filtering via Qwen2.5-VL-7B~\cite{bai2025qwen2}, resulting in a curated set of 45K high-quality grounded editing samples.

\begin{figure*}[t]
	\begin{center}
		\includegraphics[width=\linewidth]{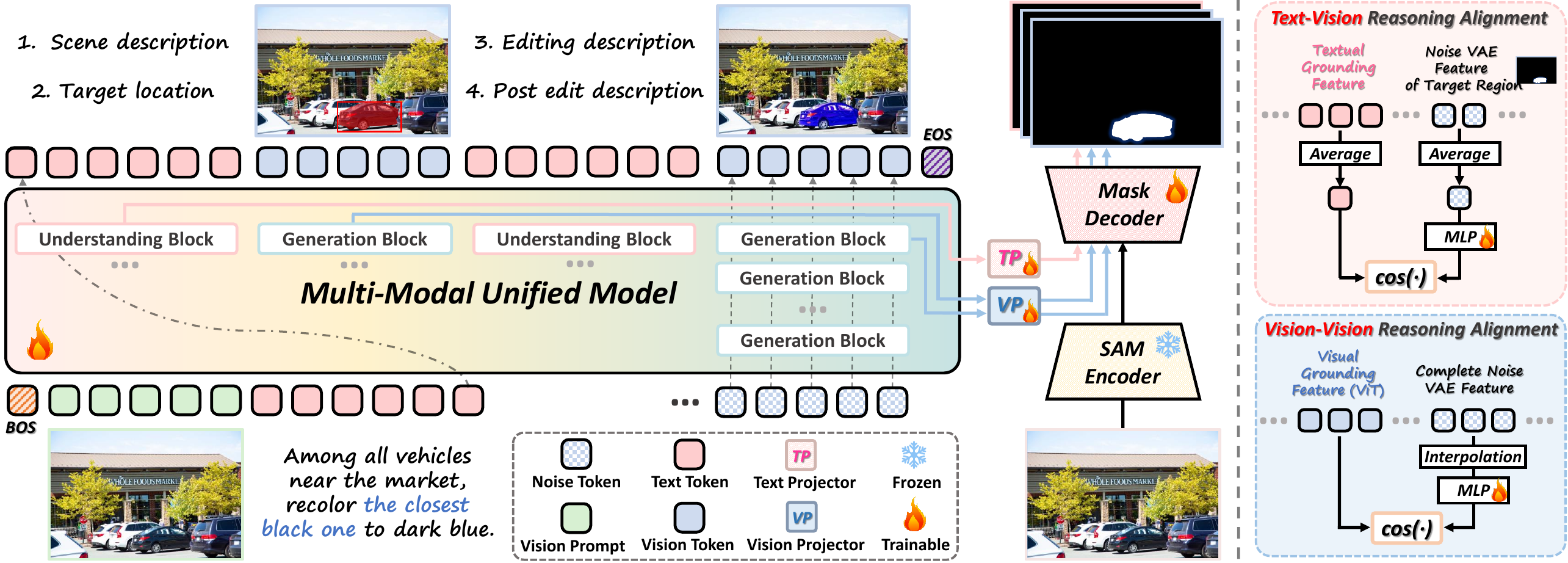}
	\end{center}
	\vspace{-6pt}
	\caption{\textbf{Overview of the Proposed InterCoG Framework.} \textbf{Left:} Our framework performs text–vision interleaved chain-of-grounding reasoning to interpret and locate user-intended targets and formulate editing descriptions, ultimately producing precise and semantically aligned editing results. \textbf{Right:} Illustration of the proposed text–vision and vision–vision reasoning alignment schemes, designed to enforce coherent multimodal grounding reasoning.}
	\label{figure3}
	\vspace{-11pt}
\end{figure*}

\subsection{Interleaved Chain-of-Grounding Framework}

Our framework utilizes a state-of-the-art unified model Bagel \cite{deng2025emerging} as the backbone, chosen for its outstanding multimodal generation capabilities. 

\noindent\textbf{Preliminaries:} Bagel comprises an understanding expert, built from understanding blocks, and a generation expert, constructed from generation blocks. The two experts share self-attention to enable cross-modal interaction and are responsible for text and vision generation, respectively. During inference, text is produced autoregressively following the next-token prediction paradigm, while images are synthesized by predicting velocity vectors within the rectified flow framework.

\noindent\textbf{Overview:} 
We build InterCoG upon the unified multimodal foundation model $\mathcal{M}$ (instantiated as Bagel~\cite{deng2025emerging}), and reformulate instruction-based image editing as a structured interleaved chain-of-grounding reasoning process. 
Given an input image $\mathbf{I} \in \mathbb{R}^{H \times W \times 3}$ and an editing instruction $\mathbf{x}$, our goal is to generate an edited image $\mathbf{I}'$ that is both semantically faithful to the instruction and spatially precise with respect to implicitly referred targets.

Instead of directly performing $\mathbf{I}' = \mathcal{M}(\mathbf{I}, \mathbf{x})$, we introduce an explicit reasoning trajectory $\mathcal{Z}$ that explicitly decomposes the editing process into three interleaved stages:
\begin{equation}
	\mathcal{Z} = \{ \mathbf{z}^{\text{text}}, \mathbf{z}^{\text{vis}}, \mathbf{z}^{\text{edit}} \},
\end{equation}
where $\mathbf{z}^{\text{text}}$, $\mathbf{z}^{\text{vis}}$, and $\mathbf{z}^{\text{edit}}$ denote textual grounding reasoning, visual grounding reasoning, and editing-oriented semantic rewriting tokens, respectively.

The overall generation process is factorized as:
\begin{equation}
	p_\theta(\mathbf{I}', \mathcal{Z} \mid \mathbf{I}, \mathbf{x})
	=
	p_\theta(\mathcal{Z} \mid \mathbf{I}, \mathbf{x})
	p_\theta(\mathbf{I}' \mid \mathbf{I}, \mathbf{x}, \mathcal{Z}),
\end{equation}
where $\theta$ denotes the parameters of the unified model.

The schematic illustration of the proposed InterCoG is depicted in Fig.~\ref{figure3}. 
Given an input image and an editing instruction, InterCoG explicitly constructs a structured reasoning trajectory through interleaved text--vision reasoning tokens within special \verb|<think></think>| segments:
\begin{equation}
	\mathbf{s} =
	[\mathbf{I}, \mathbf{x},
	\texttt{<think>},
	\mathcal{Z},
	\texttt{</think>} ].
\end{equation}

Moreover, the reasoning trajectory is progressively decomposed as:
\begin{equation}
	\begin{aligned}
		p_\theta(\mathcal{Z} \mid \mathbf{I}, \mathbf{x})
		={}&
		p_\theta(\mathbf{z}^{\text{text}} \mid \mathbf{I}, \mathbf{x}) \\
		&\cdot
		p_\theta(\mathbf{z}^{\text{vis}} \mid \mathbf{I},\mathbf{x}, \mathbf{z}^{\text{text}}) \\
		&\cdot
		p_\theta(\mathbf{z}^{\text{edit}} \mid \mathbf{I},\mathbf{x}, \mathbf{z}^{\text{text}}, \mathbf{z}^{\text{vis}}).
	\end{aligned}
\end{equation}

\noindent
\textit{(1) Textual Grounding Reasoning.}
InterCoG first generates textual grounding reasoning tokens:
\begin{equation}
	\mathbf{z}^{\text{text}}
	\sim
	p_\theta(\mathbf{z}^{\text{text}} \mid \mathbf{I}, \mathbf{x}),
\end{equation}
which resolve ambiguous references in $\mathbf{x}$ by explicitly reasoning about spatial relations. 
This stage produces structured reasoning descriptions that uniquely identify the target entity.

\noindent
\textit{(2) Visual Grounding Reasoning.}
Conditioned on the inferred textual grounding tokens, the model further generates visual grounding reasoning tokens:
\begin{equation}
	\mathbf{z}^{\text{vis}}
	\sim
	p_\theta(\mathbf{z}^{\text{vis}} \mid \mathbf{I},\mathbf{x}, \mathbf{z}^{\text{text}}),
\end{equation}
which project the inferred target into pixel space via explicit bounding boxes and segmentation masks. 
This stage bridges high-level semantic reasoning and low-level spatial grounding.

\noindent
\textit{(3) Editing-Oriented Semantic Rewriting.}
Finally, the model generates editing-oriented rewriting tokens:
\begin{equation}
	\mathbf{z}^{\text{edit}}
	\sim
	p_\theta(\mathbf{z}^{\text{edit}}
	\mid
	\mathbf{I},\mathbf{x},
	\mathbf{z}^{\text{text}},
	\mathbf{z}^{\text{vis}}),
\end{equation}
which refine the original instruction into a grounded and executable editing specification.

\noindent
\textit{(4) Final Image Generation.}
Conditioned on the complete reasoning trajectory $\mathcal{Z}$, the final edited image is generated as:
\begin{equation}
	\mathbf{I}'
	\sim
	p_\theta(\mathbf{I}' \mid \mathbf{I}, \mathbf{x},\mathcal{Z}),
\end{equation}
ensuring that edits are spatially precise and semantically faithful to the inferred target regions.

The above formulation transforms image editing from direct conditional generation into a structured reasoning-guided generation process with explicit grounding supervision. 
By interleaving $\mathbf{z}^{\text{text}}$, $\mathbf{z}^{\text{vis}}$, and $\mathbf{z}^{\text{edit}}$ within a unified sequence, InterCoG establishes a structured and interpretable grounding chain, substantially improving localization accuracy and editing faithfulness in complex multi-entity scenes.

\noindent\textbf{Multimodal Grounding Reconstruction Supervision.} 
To further enhance localization capabilities during reasoning and editing, we introduce an auxiliary grounding supervision with a mask reconstruction objective.
This supervision employs a shared mask decoder to reconstruct the binary mask conditioned on both textual and visual representations, compelling the model to capture more intrinsic grounding features and achieve more precise edits. Moreover, by sharing the decoder across modalities, the framework naturally aligns textual and visual representations, allowing the model to leverage textual reasoning to improve visual grounding and editing fidelity.

Concretely, we utilize the last hidden states from three sources:
(1) text tokens $H_t$ for textual grounding reasoning,
(2) noised VAE tokens $H_v$ for visual grounding reasoning, and
(3) noised VAE tokens $H_e$ for editing results generation. 
Each hidden state is processed by a three-layer modality-specific projector, $\mathit{TP}(\cdot)$ for text and $\mathit{VP}(\cdot)$ for vision, producing the conditioning features $C_t$, $C_v$, and $C_e$.
These features serve as conditions for a shared mask decoder, which predicts the target masks $\hat{M}_t$, $\hat{M}_v$, and $\hat{M}_e$ from the SAM-encoded image features $E$. Specifically, the text-based feature $C_t$ is averaged to yield a single embedding and incorporated through cross-attention, whereas the spatial features $C_v$ and $C_e$ are directly added to $E$, enabling spatially-aware conditioning:
\begin{equation}
	\begin{aligned} 
		\hat{M_t} &= \text{MaskDecoder}(E, C_{t}), C_t = \mathit{TP}(H_t),\\
		\hat{M_v} &= \text{MaskDecoder}(E, C_{v}), C_v = \mathit{VP}(H_v),\\
		\hat{M_e} &= \text{MaskDecoder}(E, C_{e}), C_e = \mathit{VP}(H_e).\\
	\end{aligned}
\end{equation}

The predicted masks are supervised by the ground-truth binary mask $M$ using binary cross-entropy (BCE) loss:
\begin{equation}
	\resizebox{.9\columnwidth}{!}{$
		\mathcal{L}_{mask} = \text{BCE}(\hat{M_t}, M) + \text{BCE}(\hat{M_v}, M) + \text{BCE}(\hat{M_e}, M).
		$}
\end{equation}

\noindent\textbf{Multimodal Grounding Reasoning Alignment.} Although the chain-of-grounding paradigm offers an interpretable view into the model’s intermediate grounding process, it may suffer from shortcut reasoning, resulting in discrepancies between the inferred location and the actual editing behavior.
To explicitly enforce the model to perform reasoning-consistent editing guided by multimodal grounding cues, we propose a grounding reasoning alignment scheme that regularizes the consistency between the editing features and the multimodal grounding representations.
This alignment constrains the model to execute its editing based on the regions inferred during reasoning, thereby ensuring spatially faithful and reasoning-consistent modifications.

\begin{table*}[t!]
	\centering
	\caption{Quantitative comparison on the proposed GroundEdit-Bench. EGA and ES denote the editing grounding accuracy and editing score, respectively. The “Average” column indicates the overall mean across all samples, rather than the mean of the listed categories.}
	\setlength{\tabcolsep}{1.6mm}
	\footnotesize
	\vspace{-4pt}
	\begin{tabular}{l|cc|cc|cc|cc|cc|cc|ccc}
		\toprule[1pt]
		\multirow{2}{*}{\textbf{Method}} & \multicolumn{2}{c|}{Color Change} & \multicolumn{2}{c|}{Remove} & \multicolumn{2}{c|}{Replace} & \multicolumn{2}{c|}{Add} & \multicolumn{2}{c|}{Appearance} & \multicolumn{2}{c|}{Others} & \multicolumn{3}{c}{Average}\\
		& EGA$\uparrow$ & ES$\uparrow$ & EGA$\uparrow$ & ES$\uparrow$ &  EGA$\uparrow$ & ES$\uparrow$ &  EGA$\uparrow$ & ES$\uparrow$ &  EGA$\uparrow$ & ES$\uparrow$ &  EGA$\uparrow$ & ES$\uparrow$ &  EGA$\uparrow$ & ES$\uparrow$&Human$\uparrow$\\
		\midrule
		\textcolor{gray}{FLUX.2 [Pro] \cite{flux2}} & \textcolor{gray}{0.53} & \textcolor{gray}{2.61} & \textcolor{gray}{0.57} & \textcolor{gray}{2.48} & \textcolor{gray}{0.48} & \textcolor{gray}{2.76} & \textcolor{gray}{0.61} & \textcolor{gray}{3.08} & \textcolor{gray}{0.26} & \textcolor{gray}{3.28} & \textcolor{gray}{0.38} & \textcolor{gray}{2.92} & \textcolor{gray}{0.53} & \textcolor{gray}{2.65} & \textcolor{gray}{7.4} \\
		\textcolor{gray}{GPT-Image-1 \cite{gpt-image1}} & \textcolor{gray}{0.61} & \textcolor{gray}{2.78} & \textcolor{gray}{0.74} & \textcolor{gray}{2.89} & \textcolor{gray}{0.55} & \textcolor{gray}{3.20} & \textcolor{gray}{0.42} & \textcolor{gray}{2.82} & \textcolor{gray}{0.83} & \textcolor{gray}{3.56} & \textcolor{gray}{0.65} & \textcolor{gray}{3.72} & \textcolor{gray}{0.66} & \textcolor{gray}{2.93} & \textcolor{gray}{8.4} \\
		\textcolor{gray}{GPT-Image-1.5 \cite{gpt-image1.5}} & \textcolor{gray}{0.83} & \textcolor{gray}{4.09} & \textcolor{gray}{0.84} & \textcolor{gray}{3.82} & \textcolor{gray}{0.82} & \textcolor{gray}{3.76} & \textcolor{gray}{0.80} & \textcolor{gray}{3.58} & \textcolor{gray}{0.89} & \textcolor{gray}{3.62} & \textcolor{gray}{0.83} & \textcolor{gray}{4.06} & \textcolor{gray}{0.81} & \textcolor{gray}{3.92} & \textcolor{gray}{8.7} \\
		\textcolor{gray}{Nano Banana \cite{gemini2023team}} & \textcolor{gray}{0.56} & \textcolor{gray}{2.91} & \textcolor{gray}{0.66} & \textcolor{gray}{3.15} & \textcolor{gray}{0.53} & \textcolor{gray}{2.64} & \textcolor{gray}{0.45} & \textcolor{gray}{2.58} & \textcolor{gray}{0.71} & \textcolor{gray}{4.08} & \textcolor{gray}{0.59} & \textcolor{gray}{3.40} & \textcolor{gray}{0.61} & \textcolor{gray}{2.98} & \textcolor{gray}{8.1} \\
		\textcolor{gray}{Nano Banana Pro \cite{gemini2023team}} & \textcolor{gray}{0.82} & \textcolor{gray}{4.11} & \textcolor{gray}{0.85} & \textcolor{gray}{3.94} & \textcolor{gray}{0.80} & \textcolor{gray}{3.96} & \textcolor{gray}{0.83} & \textcolor{gray}{3.72} & \textcolor{gray}{0.92} & \textcolor{gray}{3.38} & \textcolor{gray}{0.90} & \textcolor{gray}{4.30} & \textcolor{gray}{0.87} & \textcolor{gray}{4.02} & \textcolor{gray}{9.2} \\
		
		\midrule
		InstructPix2Pix \cite{brooks2023instructpix2pix} & 0.13 & 1.25 & 0.11 & 1.18 & 0.07 & 1.06 & 0.16 & 1.72 & 0.10 & 1.04 & 0.05 & 1.04 & 0.12 & 1.21 & 2.3 \\
		MagicBrush \cite{zhang2023magicbrush}& 0.23 & 1.18 & 0.15 & 1.45 & 0.13 & 1.38 & 0.12 & 1.72 & 0.13 & 1.06 & 0.32 & 2.38 & 0.19 & 1.39 & 2.1 \\
		AnyEdit \cite{yu2025anyedit}& 0.24 & 1.35 & 0.14 & 1.30 & 0.18 & 1.34 & 0.10 & 1.02 & 0.12 & 1.96 & 0.09 & 2.06 & 0.19 & 1.45 & 2.7 \\
		ICEdit \cite{zhang2025context}& 0.28 & 1.44 & 0.33 & 1.37 & 0.24 & 1.66 & 0.23 & 1.72 & 0.52 & 1.72 & 0.35 & 3.06 & 0.30 & 1.52 & 3.2 \\
		\midrule
		GoT \cite{fang2025got}& 0.45 & 2.52 & 0.56 & 2.41 & 0.49 & 3.26 & 0.32 & 2.72 & 0.78 & 3.38 & 0.60 & 2.38 & 0.52 & 2.63 & 5.8 \\
		UniWorld-v1 \cite{lin2025uniworld}& 0.34 & 1.76 & 0.39 & 1.72 & 0.40 & 2.66 & 0.54 & 2.72 & 0.18 & 2.06 & 0.22 & 2.04 & 0.37 & 1.90 & 4.5 \\
		Step1X-Edit \cite{liu2025step1x}& 0.50 & 2.15 & 0.55 & 2.21 & 0.45 & 2.56 & \underline{0.62} & \underline{3.20} & 0.66 & 3.38 & 0.50 & 3.06 & 0.55 & 2.02 & 5.0 \\
		OmniGen2 \cite{wu2025omnigen2}& 0.26 & 1.92 & 0.32 & 1.58 & 0.27 & 2.22 & 0.11 & 1.78 & 0.07 & 3.30 & 0.11 & 2.38 & 0.30 & 1.91 & 4.7 \\
		FLUX.1 Kontext [Dev] \cite{batifol2025flux}& 0.42 & 1.45 & 0.45 & 2.02 & 0.38 & 2.48 & 0.57 & 3.06 & 0.07 & 3.58 & 0.15 & \underline{3.62} & 0.42 & 1.89 & 5.2 \\
		Qwen-Image-Edit \cite{wu2025qwen}& 0.55 & 2.51 & \underline{0.65} & \underline{3.31} & 0.54 & \underline{3.44} & 0.43 & 2.56 & 0.55 & \underline{3.96} & 0.58 & 3.38 & 0.59 & \underline{2.97} & 8.3 \\
		Bagel \cite{deng2025emerging}& \underline{0.62} & \underline{2.72} & 0.65 & 2.62 & \underline{0.55} & 2.46 & 0.28 & 2.54 & \underline{0.74} & 2.84 & \underline{0.67} & 3.06 & \underline{0.64} & 2.70 & 6.9 \\
		Bagel-think \cite{deng2025emerging}& 0.58 & 2.36 & 0.55 & 2.54 & 0.43 & 2.36 & 0.23 & 1.78 & 0.64 & 3.06 & 0.46 & 3.06 & 0.56 & 2.48 & 6.5 \\
		\midrule
		Bagel-SFT & 0.76 & 2.91 & 0.71 & 3.27 & 0.65 & 3.64 & 0.63 & 3.06 & 0.84 & 3.96 & 0.74 & 3.72 & 0.72 & 3.20 & 8.6 \\
		\textbf{InterCoG (Ours)} 
		& \cellcolor{blue!10}\textbf{0.86}
		& \cellcolor{blue!10}\textbf{4.12} 
		& \cellcolor{blue!10}\textbf{0.86}
		& \cellcolor{blue!10}\textbf{3.89} 
		& \cellcolor{blue!10}\textbf{0.91}
		& \cellcolor{blue!10}\textbf{3.96} 
		& \cellcolor{blue!10}\textbf{0.88}
		& \cellcolor{blue!10}\textbf{3.80} 
		& \cellcolor{blue!10}\textbf{0.97}
		& \cellcolor{blue!10}\textbf{4.06} 
		& \cellcolor{blue!10}\textbf{0.94}
		& \cellcolor{blue!10}\textbf{4.06} 
		& \cellcolor{blue!10}\textbf{0.89}
		& \cellcolor{blue!10}\textbf{4.02}
		& \cellcolor{blue!10}\textbf{9.3} \\
		\bottomrule[1pt]
	\end{tabular}
	\label{table1}
	\vspace{-7pt}
\end{table*}

More concretely, as illustrated on the right side of Fig.~\ref{figure3}, we perform two types of alignment: text–vision reasoning alignment and vision–vision reasoning alignment. Our alignment is conducted at a specific transformer layer; hence, we omit the layer index $l$ for notational simplicity. For the text–vision reasoning alignment,
we first compute the average of the token features corresponding to \textit{“2. Target location”} in the textual grounding reasoning, denoted as $G_t^{avg}$.
In parallel, we extract the noised VAE tokens from the editing region specified by the mask and compute their average to obtain $N_r^{avg}$.
The noised representation $N_r^{avg}$ is then projected via a learnable MLP, $P_t(\cdot)$, and the resulting embedding is then aligned with the text-derived grounding cues $G_t^{avg}$ by maximizing the cosine similarity:
\begin{equation}
	\begin{split}
		G_t^{avg} = \text{Average}&(G_t), \quad N_r^{avg} = \text{Average}(N_r), \\
		\mathcal{L}_{tv} = &-\cos(P_t(N_r^{avg}), G_t^{avg}),
	\end{split}
\end{equation}
where $\cos(\cdot,\cdot)$ denotes cosine similarity.

Similarly, for vision–vision reasoning alignment, we leverage the ViT token representations \cite{deng2025emerging} of the grounded images produced by visual grounding reasoning, denoted as $G_v^{vit}$. These tokens encode richer contextual and spatial semantics than their VAE counterparts, providing a more expressive basis for aligning visual features.
To establish spatial correspondence, the noised VAE editing features $N$ are first interpolated to the spatial resolution of $G_v^{vit}$ and subsequently projected through a learnable MLP, $P_v(\cdot)$.
We then apply a cosine similarity loss to align the two representations, thereby enforcing spatially grounded coherence between the visual grounding reasoning and the editing features:
\begin{equation}
	\begin{split}
		N^r &= \text{Interp}(N, \text{size}(G_v^{vit})), \\
		\mathcal{L}_{vv} &=  -  \frac{1}{I}\sum_{i=1}^{I}  \cos(P_v(N^r), G_v^{vit}),
	\end{split}
\end{equation}
where $i$ is the patch index.
We apply the reasoning alignment loss at an intermediate transformer layer (i.e., 14), which not only provides meaningful location guidance for the subsequent layers but also allows them to focus more effectively on appearance refinement and content generation.

\noindent\textbf{Training and Inference.} 
During training, the model is optimized with a cross-entropy loss $\mathcal{L}_{ce}$ for text token prediction. For visual generation, we adopt the Rectified Flow paradigm, supervising the network to predict the velocity field along a straight-line trajectory using a mean squared error loss $\mathcal{L}_{mse}$.
In addition, the aforementioned losses $\mathcal{L}_{tv}$, $\mathcal{L}_{vv}$, and $\mathcal{L}_{mask}$ are incorporated, yielding the overall training objective:
\begin{equation}
	\mathcal{L} = \mathcal{L}_{ce} + \mathcal{L}_{mse} + \lambda_{tv} \mathcal{L}_{tv} + \lambda_{vv} \mathcal{L}_{vv} + \lambda_{mask} \mathcal{L}_{mask},
\end{equation}
where $\lambda_{tv}$, $\lambda_{vv}$, and $\lambda_{mask}$ are scaling factors, set empirically to 1.0, 1.0, and 0.1. 
During inference, the model relies solely on the main network to autoregressively generate an interleaved sequence of text and images, discarding all auxiliary modules and projection layers.

\vspace{-3pt}
\section{Experiments}

\subsection{Implementation Details}
We initialize our framework with pretrained Bagel weights and train it on the GroundEdit-45K dataset, augmented with additional general understanding, generation, and editing data \cite{byeon2022coyo,gu2022wukong,chen2025sharegpt,schuhmann2022laion,sharma2018conceptual,sun2023journeydb,schuhmann2022laion} to preserve the base model’s capabilities. The system prompt governs whether the model operates in grounding reasoning mode. Training is conducted for 4K iterations with a batch size of 64 and a learning rate of $2 \times 10^{-5}$.
To benchmark grounding-aware editing, we introduce GroundEdit-Bench, comprising 500 real-world cases with meticulously human-annotated instructions and target bounding boxes, ensuring rigorous evaluation of localization and editing accuracy.

\subsection{Comparisons with State-of-the-Arts }
\begin{figure*}[ht!]
	\begin{center}
		\includegraphics[width=\linewidth]{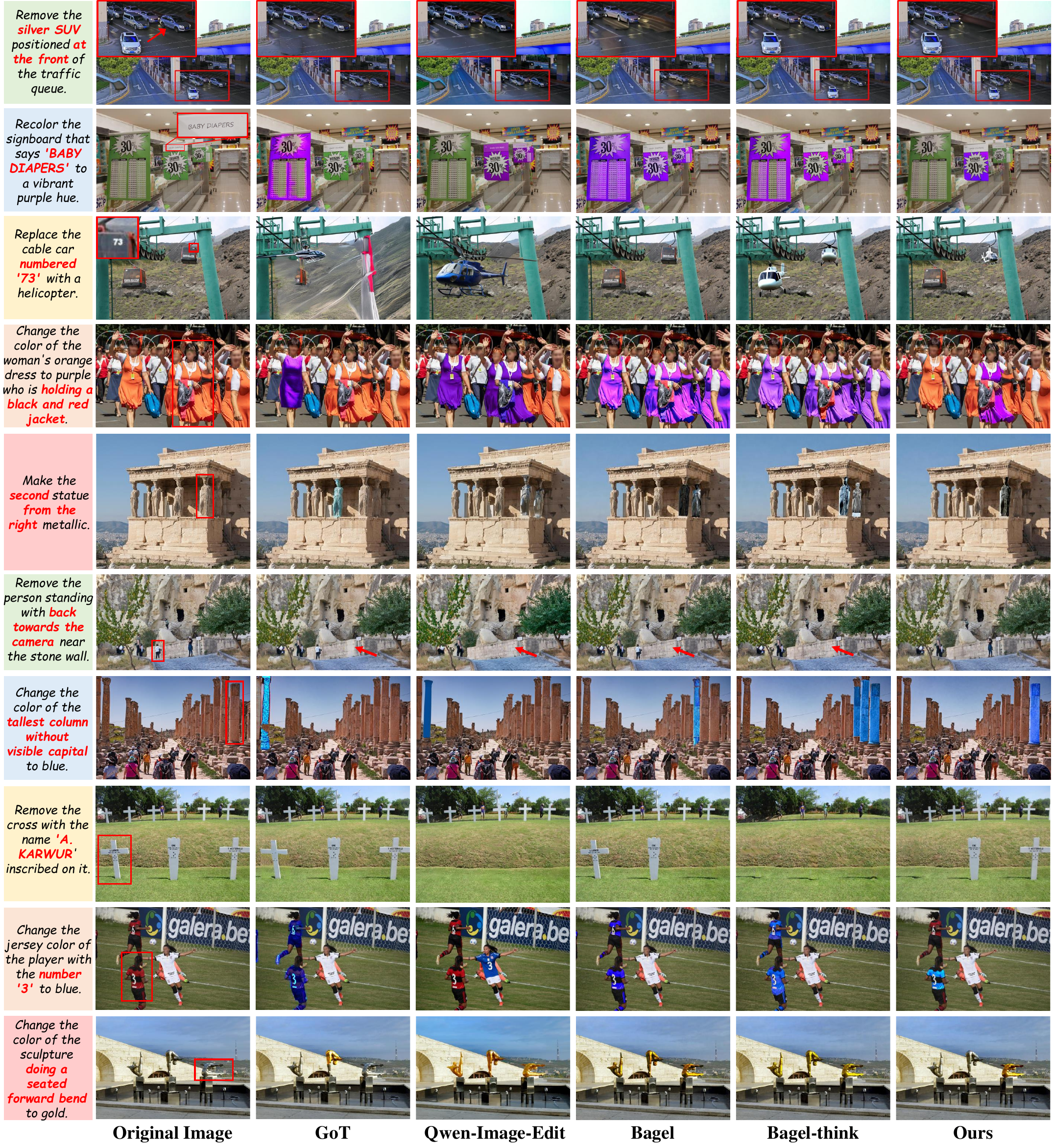}
	\end{center}
	\vspace{-10pt}
	\caption{Qualitative comparisons on our proposed GroundEdit-Bench. Our method consistently delivers highly precise and spatially accurate edits, particularly in multi-entity and fine-grained reasoning scenarios. \textbf{Best viewed at screen!} }
	\label{figure4}
	\vspace{-4pt}
\end{figure*}

\begin{figure*}[h]
	
	\begin{center}
		\includegraphics[width=\linewidth]{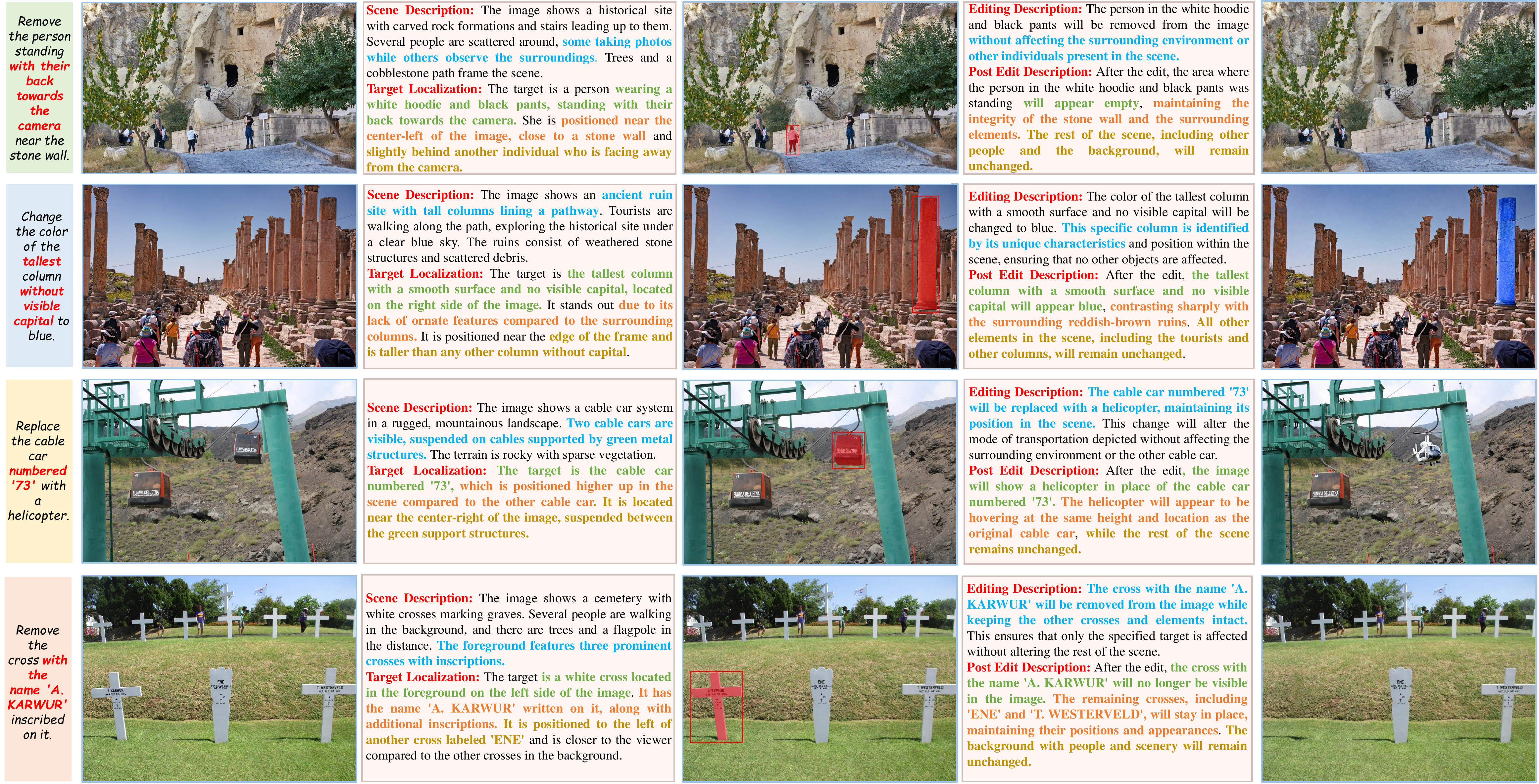}
	\end{center}
	\vspace{-10pt}
	\caption{Visualization of the interleaved chain-of-grounding reasoning processing. InterCoG first interprets user-intended referential targets via textual reasoning and then highlights the object via visualizing bounding boxes and masks in pixel space. By interleaving these multimodal grounding cues, InterCoG is able to precisely locate the editing regions and achieve spatially exact modifications. \textbf{Best viewed at screen!} }
	\label{figure5}
	\vspace{-6pt}
\end{figure*}

\noindent\textbf{Comparisons on the proposed GroundEdit-Bench.} 
To rigorously evaluate the effectiveness of InterCoG, we conduct a comprehensive comparison against state-of-the-art instruction-based editing methods, as summarized in Tab.~\ref{figure4}. 
We report two metrics: \textit{Editing Grounding Accuracy (EGA)}, which measures the proportion of the edited region (absolute difference before and after editing) that falls within the target bounding box, and \textit{Editing Score (ES)}, which evaluates overall editing quality on a 1–5 scale via GPT-4o, jointly considering visual fidelity and grounding precision. We also report mean human ratings on a 1–10 scale.
As can be found, InterCoG achieves unprecedented performance, outperforming all competitors. Notably, it surpasses the top-performing baseline Bagel \cite{deng2025emerging} by 0.25 in grounding accuracy and exceeds Qwen-Image-Edit by 1.05 in overall editing quality, demonstrating its superior ability to leverage interleaved reasoning for precise interpretation of user intention and pinpoint execution of edits.
Although GoT~\cite{fang2025got} employs textual reasoning for target localization, its reliance on MLLM-based coordinate estimation yields coarse and inaccurate grounding, while the decoupled design between grounding and generation further leads to suboptimal editing results.

We also demonstrate visual comparisons in Fig. \ref{figure4} and Fig. \ref{figure5}. As suggested, our method precisely interprets user-referred targets and performs interleaved grounding reasoning to achieve precise editing on the desired target. In contrast, existing approaches often fail to resolve referential intent, leading to unintended modifications across multiple entities.

\noindent\textbf{Comparisons on SmartEdit benchmark.}
Furthermore, we evaluate InterCoG on the reasoning scenarios of the SmartEdit benchmark \cite{huang2024smartedit} to more comprehensively validate its effectiveness. As summarized in Tab.~\ref{table_reasoning},
InterCoG delivers clear and consistent gains in PSNR, SSIM, LPIPS, and CLIP Score. These improvements confirm that our textual reasoning mechanism correctly infers user-intended targets, while the pixel-level grounding mechanism ensures highly accurate localization of the editable targets.
This interleaved design enables precise editing in complex reasoning scenarios while preserving the surrounding context with minimal disturbance. In contrast, existing methods often suffer from reasoning failures that lead to incorrect target localization, and MURE’s coarse-grained localization paradigm also results in suboptimal grounding precision.

\begin{table}[h]
	\vspace{-7pt}
	\centering
	\setlength{\tabcolsep}{2.2mm}
	\footnotesize
	\caption{Quantitative comparison results on SmartEdit benchmark (Reasoning Scenarios).}
	\vspace{-3pt}
	\begin{tabular}{l|cccc}
		\toprule[1pt]
		\multirow{2}{*}{\textbf{Method}}&\multicolumn{4}{c}{SmartEdit Reasoning Scenarios } \\
		& PSNR$\uparrow$ & SSIM$\uparrow$ & LPIPS$\downarrow$ & CLIP Score$\uparrow$ \\
		\midrule
		InstructPix2Pix  \cite{brooks2023instructpix2pix}    & 24.234 & 0.707 & 0.083 & 19.413 \\
		MagicBrush    \cite{zhang2023magicbrush}    & 22.101 & 0.694 & 0.113 & 19.755 \\
		InstructDiffusion \cite{geng2024instructdiffusion} & 21.453 & 0.666 & 0.117 & 19.523 \\
		SmartEdit-7B  \cite{huang2024smartedit}     & 25.258 & 0.742 & 0.055 & 20.950 \\
		SmartEdit-13B  \cite{huang2024smartedit}     & 25.757 & 0.747 & \underline{0.051} & 20.777 \\
			Qwen-Image-Edit \cite{wu2025qwen}	&21.772	&0.657	&0.084	&21.012\\
		Bagel       \cite{deng2025emerging}         & 28.076 & 0.839 & 0.060 & 20.767 \\
		MURE    \cite{zou2025beyond}             & \underline{28.694} & \underline{0.883} & 0.062 & \underline{21.298} \\
		\textbf{InterCoG (Ours)} &\textbf{30.024}&\textbf{0.901}&\textbf{0.045}&\textbf{21.314}\\
		\bottomrule[1pt]
	\end{tabular}
	\label{table_reasoning}
	\vspace{-4pt}
\end{table}

Beyond the reasoning scenarios, we also evaluate our method on the understanding scenarios of the SmartEdit benchmark, as shown in Tab. \ref{table_reasoning2}. It is observed that our method still achieves the best performance on PSNR, SSIM, and LPIPS metrics. In contrast, the concurrent MURE model failed to fully leverage the pixel-level localization capabilities of visual reasoning and neglected the consistency between intermediate reasoning and the final results, leading to suboptimal localization editing performance in understanding scenarios.

\begin{table}[h]
	\vspace{-4pt}
	\centering
	\setlength{\tabcolsep}{2.2mm}
	\footnotesize
	\caption{Quantitative comparison results on SmartEdit benchmark (Understanding Scenarios).}
	\vspace{-4pt}
	\begin{tabular}{l|cccc}
		\toprule[1pt]
		\multirow{2}{*}{\textbf{Method}}&\multicolumn{4}{c}{SmartEdit Understanding Scenarios } \\
		& PSNR$\uparrow$ & SSIM$\uparrow$ & LPIPS$\downarrow$ & CLIP Score$\uparrow$ \\
		\midrule
		InstructPix2Pix \cite{brooks2023instructpix2pix}       & 21.576 & 0.721 & 0.089 & 22.762 \\
		MagicBrush  \cite{zhang2023magicbrush}     & 18.120 & 0.680 & 0.143 & 22.620 \\
		InstructDiffusion \cite{geng2024instructdiffusion} & 23.258 & 0.743 & 0.067 & 23.080 \\
		SmartEdit-7B  \cite{huang2024smartedit}    & 22.049 & 0.731 & 0.087 & 23.611 \\
		SmartEdit-13B   \cite{huang2024smartedit}   & 23.596 & 0.751 & 0.068 & 23.536 \\
		Qwen-Image-Edit	\cite{wu2025qwen} &22.274	&0.741	&0.086	&23.879\\
		Bagel        \cite{deng2025emerging}      & 23.823 & 0.892 & 0.083 & 23.842 \\
		MURE        \cite{zou2025beyond}         & \underline{25.611} & \underline{0.897} & \underline{0.065} & \underline{23.947} \\
		\textbf{InterCoG (Ours)} &\textbf{25.774}&\textbf{0.905}&\textbf{0.062}&\textbf{24.026}\\
		\bottomrule[1pt]
	\end{tabular}
	\label{table_reasoning2}
	\vspace{-4pt}
\end{table}

\noindent\textbf{Comparisons on GEdit-Bench.}
Furthermore, we also extend our evaluation to the widely recognized GEdit-Bench \cite{liu2025step1x}.
As summarized in Tab.~\ref{table2}, although our training data is derived from Bagel, the model trained under our interleaved text–vision reasoning paradigm surpasses Bagel itself in overall editing quality.
Such results provide compelling evidence that our framework not only facilitates grounding-oriented editing but also preserves strong general editing competence, thereby validating the generality and effectiveness of the InterCoG paradigm.
\begin{table}[h]
	\centering
	\tabcolsep=17pt
	\makeatletter\def\@captype{table}\makeatother
	\caption{Comparison of Semantic Consistency (SC), Perceptual Quality (PQ), and Over all Score (O) on GEdit-Bench.}
	\vspace{-4pt}
	\footnotesize
	\begin{tabular}{l|ccc}
		\toprule[1pt]
		\multirow{2}{*}{\textbf{Method}}&\multicolumn{3}{c}{GEdit-Bench-EN }\\
		& SC  $\uparrow$ & PQ  $\uparrow$ & O  $\uparrow$\\
		\midrule
		InstructPix2Pix \cite{brooks2023instructpix2pix}&3.58 &5.49&3.68\\
		MagicBrush \cite{zhang2023magicbrush}  &4.68 &5.66&4.52\\
		AnyEdit \cite{yu2025anyedit}&3.18 &5.82&3.21\\
		UniWorld-v1 \cite{lin2025uniworld} &4.93& 7.43& 4.85\\
		OmniGen2  \cite{wu2025omnigen2}&7.16& 6.77& 6.41\\
		Bagel  \cite{deng2025emerging}  &\underline{7.36}& \textbf{6.83}& \underline{6.52}\\
		\textbf{InterCoG (Ours))} &\textbf{7.53}&\underline{6.81}&\textbf{6.63}\\
		\bottomrule[1pt]
	\end{tabular}
	\label{table2}
	\vspace{-4pt}
\end{table}

\begin{figure*}[t]
	\vspace{-4pt}
	\begin{center}
		\includegraphics[width=\linewidth]{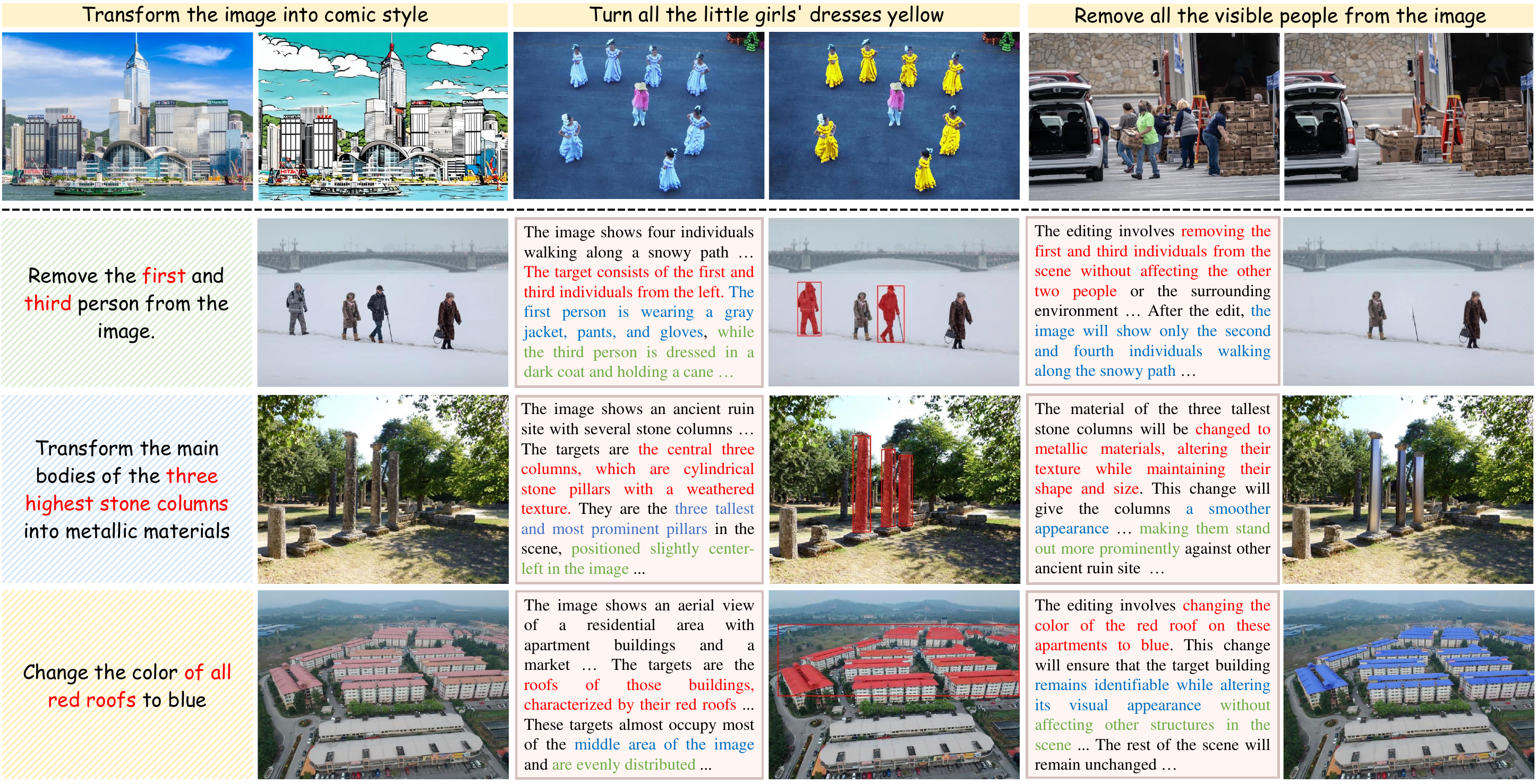}
	\end{center}
	\vspace{-4pt}
	\caption{ Visualization of model performance on lobal and multi-object editing. }
	\label{figure_multi}
	\vspace{-4pt}
\end{figure*}

\noindent\textbf{Generalization to global and multi-entry conditions.} 
Our approach naturally generalize to multi-object editing scenarios. As illustrated in Fig.~\ref{figure_multi}, the proposed chain-of-grounding paradigm precisely localizes each target via both textual grounding and visual grounding, enabling fine-grained edits on multiple entities while preserving irrelevant content. In contrast, existing methods such as Bagel and Qwen-Image-Edit are typically unable to identify the specified targets, thus failing to faithfully follow the editing instruction.

\subsection{Ablation Studies}
In Tab. \ref{table_pc}, we conduct ablation experiments on the components introduced in InterCoG. The effectiveness of each component in InterCoG is evaluated by excluding it from the model, revealing their individual contributions to the overall performance. More detailed analyses are as below.

\begin{figure}[h]
	\vspace{-8pt}
	\begin{center}
		\includegraphics[width=\linewidth]{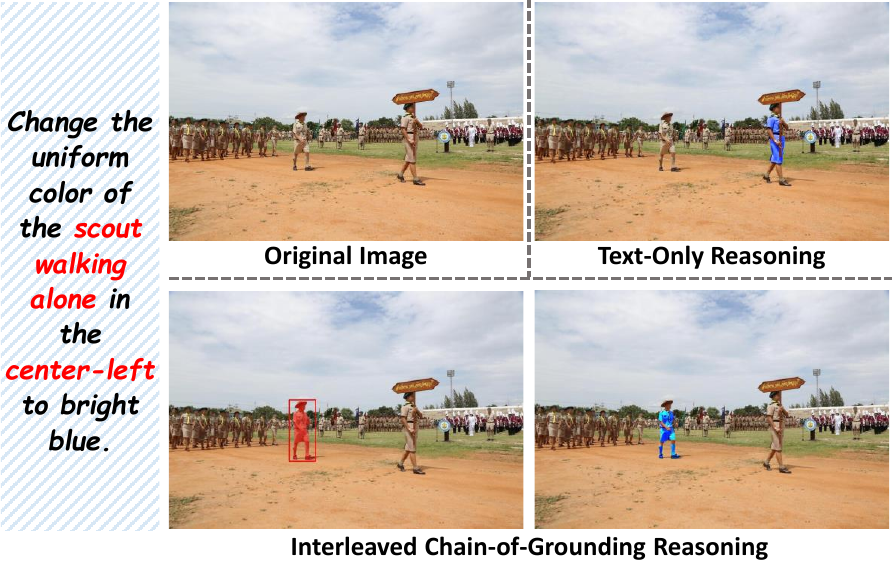}
	\end{center}
	\vspace{-8pt}
	\caption{ Visual comparison between text-only grounding reasoning and interleaved chain-of-grounding reasoning. }
	\label{figure6}
	\vspace{-4pt}
\end{figure}

\begin{table}[h]
	\vspace{-7pt}	
	\caption{Ablation studies on the GroundEdit-Bench.}
	\tabcolsep=15pt
	\vspace{-7pt}
	\footnotesize
	\begin{center}
		\begin{tabular}{l|cc}
			\toprule[1pt]
			\textbf{Variant}&EGA$\uparrow$&ES$\uparrow$\\
			\midrule
			w/o Reasoning &0.70&3.15\\
			w Text-Only Grounding Reasoning &0.76&3.57\\
			w Vision-Only Grounding Reasoning &0.74&3.23\\
			\midrule
			w/o Grounding Reconstruction &0.85&3.83\\
			w/o Reasoning Alignment &0.83&3.71
			\\
			\midrule
			InterCoG (Ours) &0.89&4.02\\
			\bottomrule[1pt]
		\end{tabular}
	\end{center}
	\label{table_pc}
	\vspace{-10pt}
\end{table} 
\noindent\textbf{Effect of Interleaved Chain-of-Grounding Reasoning.} 
As shown in Tab.~\ref{table_pc} and Fig.~\ref{figure6}, relying solely on textual grounding reasoning allows the model to parse user-referred targets but suffers from coarse positional granularity, often failing to accurately localize the intended editing regions and producing suboptimal results. In contrast, using visual grounding reasoning alone provides precise target localization but lacks the capacity for deeper referential reasoning.
The proposed interleaved grounding reasoning overcomes both limitations by jointly performing referential reasoning and supplying fine-grained positional cues. Moreover, directly generating grounding-labeled images further reinforces the model’s awareness of where to edit, substantially improving the precision of grounding-oriented editing.

\begin{figure}[h]
	\vspace{-8pt}
	\begin{center}
		\includegraphics[width=\linewidth]{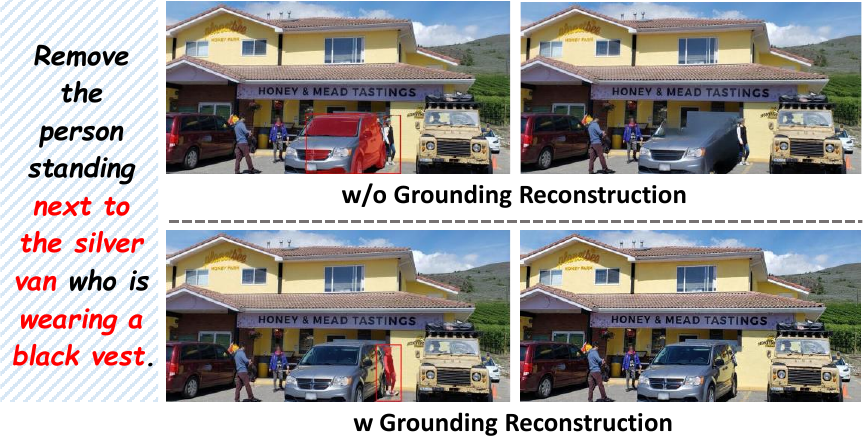}
	\end{center}
	\vspace{-8pt}
	\caption{ Visual comparison with or without our proposed multimodal grounding reconstruction supervision. }
	\label{figure8}
	\vspace{-4pt}
\end{figure}

\noindent\textbf{Effect of Multimodal Grounding Reconstruction.} 
Introducing mask decoding as an auxiliary supervision not only drives the generation branch to learn more profound grounding-aware visual representations but also enhances cross-modal alignment via a shared decoder.
As shown in Tab.~\ref{table_pc} and Fig.~\ref{figure8}, this component facilitates more robust and region-aware grounding, ultimately leading to improved localization performance.

\begin{figure}[h]
	\vspace{-8pt}
	\begin{center}
		\includegraphics[width=\linewidth]{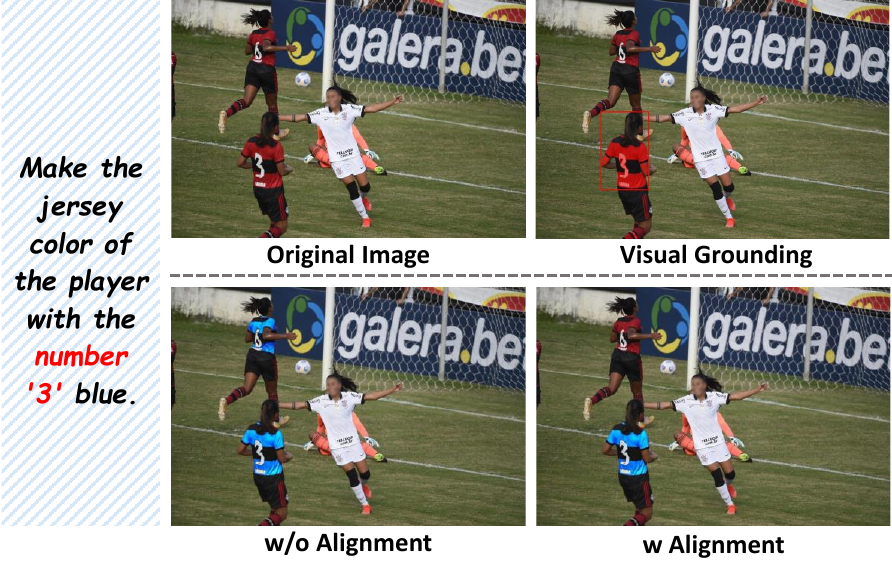}
	\end{center}
	\vspace{-8pt}
	\caption{ Visual comparison with or without our proposed multimodal grounding reasoning alignment. }
	\label{figure7}
	\vspace{-4pt}
\end{figure}

\noindent\textbf{Effect of Multimodal Grounding Reasoning Alignment.} 
As illustrated in Fig.~\ref{figure7}, without reasoning alignment, the model tends to take shortcuts, failing to fully leverage the intermediate grounding cues, which leads to discrepancies between the reasoning process and the final editing outcome.
In contrast, equipping the model with reasoning alignment enforces consistency between its reasoning trajectory and the final spatial representation, thereby realizing \textit{“what is thought is what is done.”}

\noindent\textbf{Ablation of the reasoning alignment position.}
We further conduct experiments to examine how applying reasoning alignment at different transformer layers affects performance. From Tab. \ref{table_pc1}, we observe that mid-layer alignment consistently yields the best results. In contrast, performing alignment at very shallow layers provides insufficient constraint, while applying it at very deep layers interferes with the model’s ability to learn appearance refinement and content generation.

\begin{table}[h]
	\vspace{-3pt}	
	\caption{Ablation studies of layer choices for multimodal grounding reasoning alignment on GroundEdit-Bench.}
	\tabcolsep=33pt
	\vspace{-6pt}
	\footnotesize
	\begin{center}
		\begin{tabular}{l|cc}
			\toprule[1pt]
			\textbf{Layer}&EGA$\uparrow$&ES$\uparrow$\\
			\midrule
			3 &0.79&3.70\\
			10 &0.83&3.95\\
			14 &0.89&4.02\\
			28 &0.84&3.89\\
			\bottomrule[1pt]
		\end{tabular}
	\end{center}
	\label{table_pc1}
	\vspace{-24pt}
\end{table}

\subsection{Inference Complexity.} Test-time scaling techniques inherently incur additional computational overhead versus non-reasoning models. As shown in Tab.~\ref{table_inference}, the chain-of-grounding paradigm introduces extra latency relative to the standard editing pipeline; nevertheless, this overhead remains acceptable given the performance gains. Our method is 26s slower than Qwen-Image-Edit, yet improves EGA and ES by 0.30 and 1.05, while using substantially less peak GPU memory.

\begin{table}[h]
	\centering
	\vspace{-3pt}
	\tabcolsep=5.5pt
	\makeatletter\def\@captype{table}\makeatother
	\caption{Average latency and peak GPU memary comparison under 1024$\times$1024 resolution on an NVIDIA A100 GPU.}
	\vspace{-6pt}
	\footnotesize
	\begin{tabular}{l|cccc}
		\toprule[1pt]
		\textbf{Method}&Latency&Peak GPU Mem.&EGA&ES\\
		\midrule
		Qwen-Image-Edit&1m 50s&60.6GB&0.59&2.97\\
		Bagel  &1m 03s&30.6GB&0.64&2.70\\
		InterCoG w/o Reasoning &1m 03s&30.6GB&0.75&3.51\\
		InterCoG (Ours) &2m 16s&31.9GB&0.89&4.02\\
		\bottomrule[1pt]
	\end{tabular}
	\label{table_inference}
	\vspace{-4pt}
\end{table}

It is worth noting that the additional overhead mainly affects latency, while peak GPU memory remains manageable, as most memory is consumed by model parameters rather than intermediate activations. Consequently, interleaved chain-of-grounding reasoning is not memory-constrained and does not require substantially more GPU resources for deployment on real devices. Furthermore, we observe that even when reasoning is disabled during inference, training with the proposed reasoning paradigm still yields significant performance improvements, resulting in a “free” gain. This is primarily because our training framework reinforced the model's intrinsic localization reasoning capabilities as well as modal consistency between generation and understanding, which should further demonstrate the effectiveness of our method.

\subsection{Failure Cases}
\begin{figure}[h]
	\vspace{-4pt}
	\begin{center}
		\includegraphics[width=\linewidth]{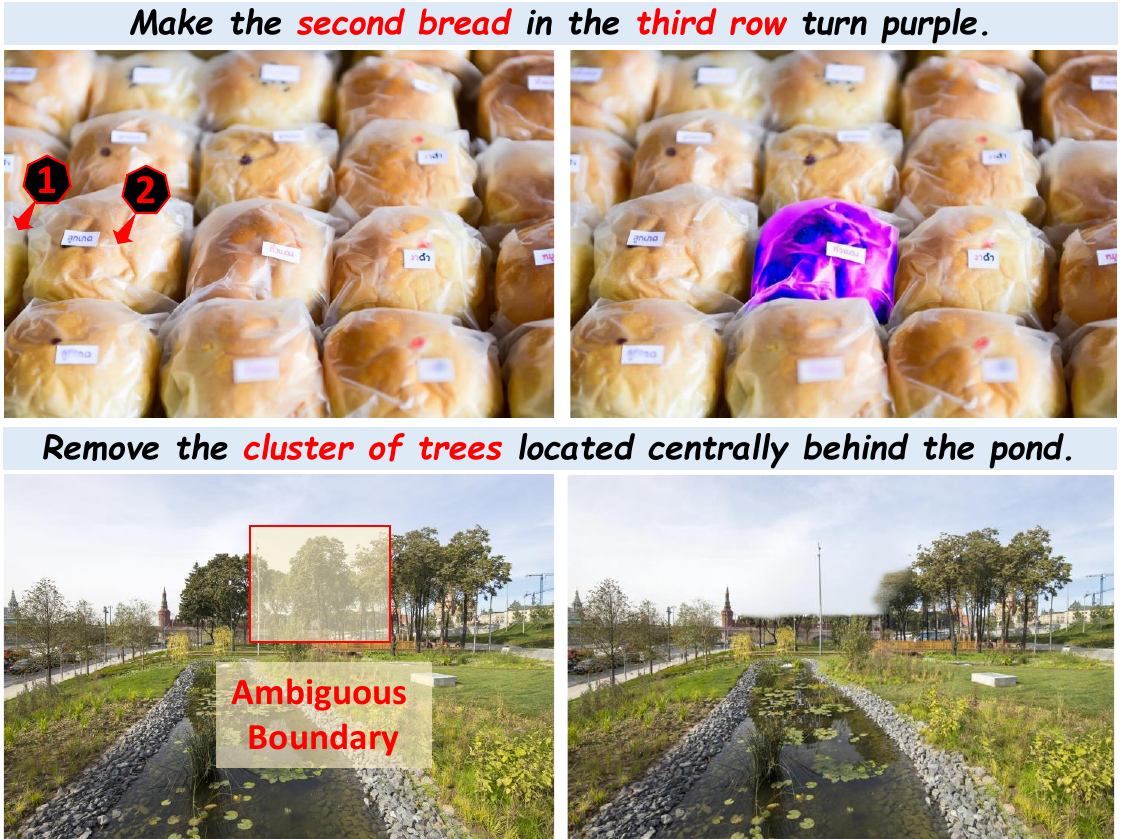}
	\end{center}
	\vspace{-4pt}
	\caption{Failure cases with ambiguous position and boundary. }
	\label{figure_failure}
	\vspace{-10pt}
\end{figure}

In Fig.~\ref{figure_failure}, we report two failure cases of our method when facing ambiguous position definition and uncertain object boundaries. The first arises in incomplete instance ordering scenarios, where our method tends to prioritize the first fully visible instance while overlooking severely occluded or fragmented ones. The second case involves editing targets with uncertain or fuzzy boundaries; when users provide ambiguous regions, our method may produce unclear boundary delineation, particularly around junctions or adjacent structures. For future work, we plan to further investigate the inference and modeling of ambiguous user intent, enabling more robust handling of weakly constrained editing instructions and yielding edits that better align with user expectations.

\vspace{-3pt}
\section{Concluding Remarks}
In this work, we present InterCoG, an interleaved chain-of-grounding reasoning framework that unifies multimodal reasoning and spatial grounding to enable precise fine-grained image editing in complex multi-subjects scenes.
InterCoG introduces an interleaved grounding paradigm that first interprets user intent through textual reasoning and then performs pixel-level visual grounding, effectively bridging high-level semantics and low-level spatial precision for coherent and controllable edits.
To further advance this paradigm, we developed GroundEdit-45K and GroundEdit-Bench, a paired dataset and benchmark designed for training and evaluating grounding-oriented editing under complex real-world conditions. Extensive experiments validate the superior reasoning coherence and grounding precision of InterCoG, highlighting its potential applications udner complex real-world editing systems.
We expect this work to provide insights into more challenging editing conditions and steer future research on this Gordian knot.

\bibliographystyle{IEEEtran}
\bibliography{egbib}

\end{document}